\pdfoutput=1

\documentclass[11pt]{article}

\usepackage[preprint]{acl}

\usepackage{times}
\usepackage{latexsym}
\usepackage{adjustbox}
\DeclareMathAlphabet\mathbfcal{OMS}{cmsy}{b}{n}
\usepackage{amssymb}
\usepackage{amsmath}
\usepackage{graphicx}
\usepackage{multirow,booktabs}
\usepackage{xcolor}
\usepackage{CJKutf8}
\usepackage{amsthm}
\usepackage{mathrsfs}
\graphicspath{{Images/}}
\usepackage[T1]{fontenc}

\usepackage[utf8]{inputenc}

\usepackage{microtype}

\usepackage{inconsolata}

\title{Personality-aware Human-centric Multimodal Reasoning: A New Task, Dataset and Baselines}

\author{Yaochen Zhu, Xiangqing Shen, and Rui Xia \\
        School of Computer Science and Engineering, \\ Nanjing University of Science and Technology, China \\
        \texttt{\{yczhu, xiangqing.shen, rxia\}@njust.edu.cn}}

\begin{document}
\maketitle

\begin{figure*}[htb]
    \centering
    \includegraphics[width=\textwidth]{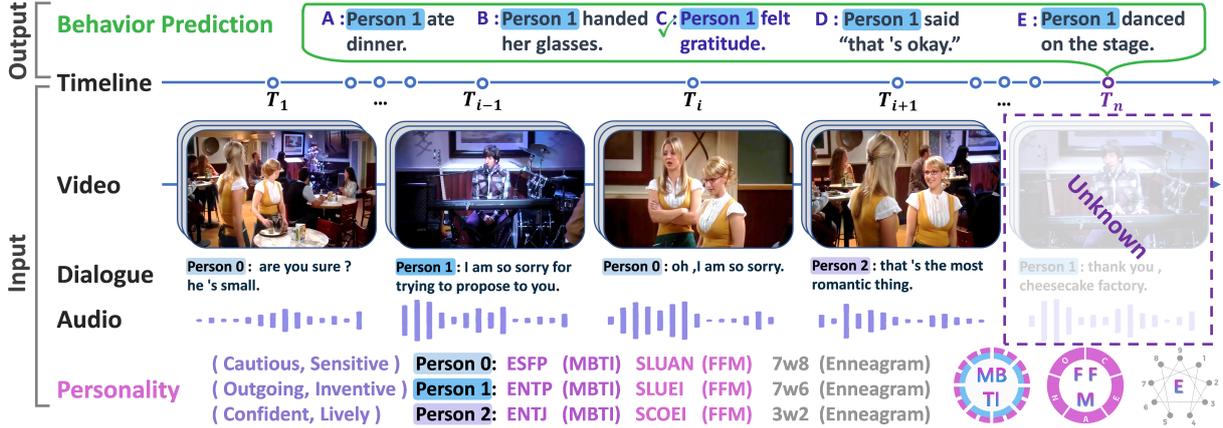}
    \caption{
    Illustrations of PHMR and PHMRD.
    Person $\mathit{0}$, $\mathit{1}$, and $\mathit{2}$ represent the characters appearing in the video clip.
    Our task is to predict the most plausible behavior description at a \textcolor[RGB]{112,48,160}{$T_n$} based on video, dialogue, audio and personalities information.
    }  
    \label{fig:T1}
\end{figure*}
\begin{abstract}
Personality traits, emotions, and beliefs shape individuals' behavioral choices and decision-making processes. However, for one thing, the affective computing community normally focused on predicting personality traits but overlooks their application in behavior prediction.
For another, the multimodal reasoning task emphasized the prediction of future states and behaviors but often neglected the incorporation of individual personality traits.
In this work, we introduce a new task called Personality-aware Human-centric Multimodal Reasoning (PHMR) ($T^1$), with the goal of forecasting the future behavior of a particular individual using multimodal information from past instances, while integrating personality factors.
We accordingly construct a new dataset based on six television shows, encompassing 225 characters and 12k samples.
To establish a benchmark for the task, we propose seven baseline methods: three adapted from related tasks, two pre-trained model, and two multimodal large language models.
The experimental results demonstrate that incorporating personality traits enhances human-centric multimodal reasoning performance.
To further solve the lack of personality annotation in real-life scenes, we introduce an extension task called Personality-predicted Human-centric Multimodal Reasoning task ($T^2$) along with the corresponding dataset and method.
We will make our dataset and code available on GitHub. \footnote{We have presented 10 examples of PHMRD in \href{https://anonymous. 4open. science/r/PHMRD_case-5962/README. md}{anonymous GitHub}. }

\end{abstract}

\section{Introduction}

An individual's personality traits, emotions, and beliefs are essential components of their individual differences, and they shape their behavioral choices and decision-making processes in different situations. 
Personality encapsulates an individual's psychological characteristics and behavioral tendencies. 
The reactions of individuals, characterized by diverse personality traits, to the identical situations can exhibit significant variations. 
For example, when faced with an event of ``Person $\mathit{1}$ sing a song in public for Person $\mathit{0}$ '', a \emph{cautious} and \emph{sensitive} Person $\mathit{0}$ may feel embarrassed, whereas a \emph{confident} and \emph{lively} Person $\mathit{0}$ may feel excited. 

However, on the one hand,
in the field of affective computing, there is often a focus on predicting and analyzing personality traits, without applying them to the prediction of future events related to individuals. 
On the other hand,recently a category of human-centric multimodal reasoning tasks, including Social-IQ~\citep{social_qa}, VLEP~\citep{VLEP}, and PMR~\citep{PMR}, has emerged with the aim to infer individuals' psychological states and behaviors by leveraging the available multimodal information.
Existing studies primarily concentrated on objective information related to individuals, while neglecting the incorporation of individual personality traits in reasoning. 

In this work, we introduce a new task called Personality-aware Human-centric Multimodal Reasoning (PHMR) ($T^1$). 
The goal of our task is to forecast the most probable behavior of a particular individual in future scenarios within intricate social interactions that encompass multiple individuals and long-term interactions. 
This is achieved by integrating multimodal signals from past moments and considering their personality traits. 

To support this task, we construct a dataset based on six television shows from TVQA~\cite{tvqa}, which we refer to as the Personality-aware Human-centric Multimodal Reasoning Dataset~(PHMRD). 
We acquire the personality of the main characters through the PDB\footnote{\url{https://www.personality-database. com}} website,
and utilize them as the personality annotations in our dataset. 
Following the mainstream settings of multimodal reasoning tasks, we also define our task as a multiple choice problem. 
As shown in Fig.~\ref{fig:T1}, Person $\mathit{0}$, $\mathit{1}$, and $\mathit{2}$ represent different characters within the video clip. 
Given the following information:
1)~the image frame sequence of the video;
2)~the utterances of dialogue;
3)~the audio waveform of the video;
4)~the personalities of the corresponding characters. 
Our task aims to select an option that best describes the most possible behavior of Person $\mathit{0}$ in the future. 
For example, it should be inferred that ``Person $\mathit{0}$ felt gratitude'' is the most possible option. 

We benchmark the PHMR task by designing seven baseline methods: three adapted from related tasks, two pre-trained multimodal model, and two multimodal large language models. 
We employ a pre-trained multimodal model to extract multimodal features and incorporate personality traits, to predict the most possible behavior. 
The personality is represented in the form of embeddings and are effectively integrated with other multimodal features. 

The experimental results reveal that the incorporation of personality traits can enhance reasoning performance in both unimodal and multimodal settings. 
Our proposed approach outperforms the adapted baselines, demonstrating improved performance. 
Our multimodal ablation experiments reveal that incorporating personality can yield performance improvements in unimodal setting for PHMR.

To further solve the lack of personality annotation in real-life scenes, we introduce an additional task called Personality-predicted Human-centric Multimodal Reasoning ($T^2$) along with the corresponding dataset. 
Taking into account the notion that an individual's personality can be inferred from their behavior and speech, we commence by predicting their personality using multimodal information.
Subsequently, we leverage these predictions as surrogates for personality annotations to enhance the reasoning process ($T^1$). 
The experimental results show that our method can effectively predict these personality, and achieves satisfactory multimodal reasoning performance without relying on personality annotations. 

\section{Task} 
\label{sec:task}
We introduce our primary task as Personality-aware Human-centric Multimodal Reasoning ($T^1$). 
Our focus lies on scenarios involving multiple individuals in long-term interactions, where each person's behavior is influenced by not only the actions of others, but also their own personality traits. 

Specifically, four types of information are provided for a video clip ${\mathcal{VC}}$:
\begin{enumerate}
\setlength{\itemsep}{0pt}
\setlength{\parskip}{0pt}
\setlength{\parsep}{0pt}
 \item \textbf{${\rm Video\,(\mathcal{V})}$}: The image frame sequences of the video. 
 \item \textbf{${\rm Audio\,(\mathcal{A})}$}: The audio waveforms of the video. 
 \item \textbf{${\rm Dialogue\,(\mathcal{D})}$}: Textual utterances of dialogue, character, and corresponding time span $T^D$. 
 \item \textbf{${\rm Personality\,(\mathcal{P})}$}: Annotated personality for characters appearing in ${\mathcal{VC}}$. 
\end{enumerate}
For a fragment ${\mathcal{VC}}$, it can be divided into $n$ time segments $T_1 \sim T_n$. 
We remove all multimodal information at the moment $T_n$. 
Given the personalities ${\mathcal{P}}$ and multimodal information $\mathcal{V}$, $\mathcal{A}$, $\mathcal{D}$ at the $T_1 \sim T_{n-1}$, the purpose of this task is to predict the most probable behavior $\mathcal{B}$ of a specific individual at time $T_n$ . 
This can be expressed as follows:
\begin{equation} 
\mathcal{B} = {\rm PHMR}(\mathcal{D},\, \mathcal{V},\,\mathcal{A},\, \mathcal{P}) ,
\end{equation}
where ${\rm PHMR}$ represents the Personality-aware Human-centric Multimodal Reasoning task. 

We model PHMR as a multiple choice task, following the established settings of other multimodal reasoning tasks~\citep{social_qa,VLEP,PMR}, where the objective is to select the most probable answer from a set of five options, denoted as \textbf{${\rm Choices\,(\mathcal{C})}$}. 

\section{Dataset}

\begin{table}
\centering
\begin{adjustbox}{max width=0.8\columnwidth}
\begin{tabular}{*{5}{lc}}
\toprule
{} &{Train} &{Dev}& {Test} & {Total}\\
\midrule
{\# Sample} & {8768} &{1938} &{1910} &{12616} \\
{Video (s)} & {74.67} &{75.84} &{75.32} &{74.95} \\
\midrule
{\# Dialogue} & {15.57} &{15.46} &{15.07} &{15.48} \\
{\# Personality} & {3.72} &{3.81} &{3.74} &{3.73} \\
\midrule
{avg. Choice}& {15.24} &{15.40} &{15.21} &{15.26} \\
{avg. Dialogue} & {6.79} &{6.60} &{6.58} &{6.73} \\
\bottomrule
\end{tabular}
\end{adjustbox}
\caption{Statistic of our PHMRD dataset. avg. represent average word of.
}
\label{tab:Statistics_of_task1}
\end{table}
\label{sec:Dataset_Construction}
\subsection{Dataset Source}

We choose TVQA~\citep{tvqa} as the data source. 
TVQA is a large-scale video question answering dataset based on television shows, encompassing QA pairs related to object relation and human activities. 
Our task emphasizes human-centric question answering and primarily concerns the behavior and psychology of individuals. 

\subsection{Dataset Construction Process}
\paragraph{Data Filtering}
To make the data more suitable for our task settings, we use ChatGPT (gpt-3.5-turbo)
to exclude QA pairs that are not related to the individual's behavior and psychology. 
In the given instance, we consider the query ``What color is the object?'' to be unrelated to individuals, while the inquiry ``Why does person $\mathit{0}$ like this object?'' is regarded as pertinent to human subjects. 
The prompt and additional examples are provided in Appendix~\ref{sec:example}. 
The above example can be accurately evaluated by ChatGPT. 
To evaluate the results of ChatGPT, we manually label two hundred samples. 
Subsequently, a kappa test is performed on the manually labeled samples and results obtained from ChatGPT. 
The kappa result is $0. 89$, indicating a strong consistency between ChatGPT and the manually labeled samples. 

\paragraph{Rewriting of Behavior Description}

We model the multimodal reasoning task as a multiple choice problem, following the mainstream setting of multimodal reasoning tasks (Social-IQ, VLEP, PMR). 
The correct option in PHMRD is derived from the rewriting of correct option in TVQA, and the process for obtaining the incorrect options followed the same methodology. 
In our preliminary experiment, it was observed that when the options are presented in the form of QA pairs, the model tends to over emphasize the importance of questions.
Therefore, 
we manually rewrite its QA pairs by converting them into declarative sentences to describe the behaviors. 
For example, given the question in TVQA: ``What does the person $\mathit{0}$ felt when Person $\mathit{1}$ sang a song in public?'',
and the answer: ``Person $\mathit{0}$ felt embarrassed. '', the rewritten result is ``Person $\mathit{0}$ felt embarrassed when Person $\mathit{1}$ sang a song in public''. 
An example of PHMRD is presented in Fig.~\ref{fig:T1}. 
We utilize ChatGPT to rewrite the behavior description according to the aforementioned settings. 
We select 500 samples and manually perform the rewriting process to obtain the manually rewriting the behavior description.
To measure the inter-annotator agreement, we utilize the MASI\footnote{A measure of the distance between collections used to quantify the degree of overlap between annotations of collection data. }~\citep{MASI} metric, and the resulting score is 0.72. 

\begin{figure*}[!htb]
    \centering
    \includegraphics[width=\textwidth]{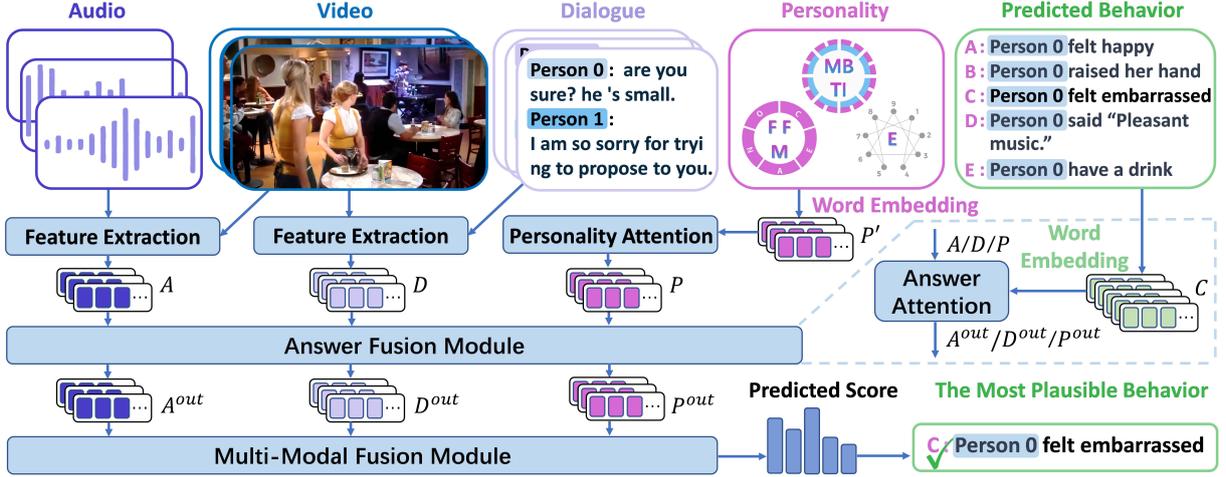}
    \caption{
    PRM$_{pretrain}$ model for PHMR.
    }
    \label{fig:model}
\end{figure*}
\begin{table}
\centering
\small
\begin{adjustbox}{max width=0.8\columnwidth}
\begin{tabular}{*{5}{lc}}
\toprule
{\textbf{TV Show}} &{Train} &{Dev}& {Test} & {Total}\\
\midrule
{\# Friends}    & {2602} &{559} &{558} & {3719}\\
{\# BBT}        & {2548} &{549} &{553} &{3650}\\
{\# Castle}     & {1525} &{329} &{333} &{2187}\\
{\# Met}        & {843} &{214} &{190} &{1247}\\
{\# House}      & {745} &{167} &{165} &{1077}\\
{\# Grey}       & {505} &{120} &{111} &{736}\\
\midrule
{Sum} & {8768} &{1938} &{1910} &{12616}\\
\bottomrule
\end{tabular}
\end{adjustbox}
\caption{Statistics regarding the sample numbers of six television shows. BBT, Met, House, and Grey represent The Big Bang Theory, How I Met You Mother, House M.D. and Grey, respectively.
}
\label{tab:six_tv_shows}
\end{table}
\paragraph{Personality Annotation}
\begin{table}
\centering
\begin{adjustbox}{max width=0.8\columnwidth}
\begin{tabular}{cc}
\toprule
{\textbf{Personality}} &{\textbf{Vocabulary}}  \\
\midrule
{$\mathtt{FFM}$} & 
{$\mathtt{[RLOAI, SCUEN, \cdots, RCUEI, SLOAN]}$
} \\
{$\mathtt{MBTI}$} & {$\mathtt{
[INTJ, ESFP, \cdots, ENFJ, ISTP]}$
} \\
{$\mathtt{Enneagram}$} & {$\mathtt{
[1w2, 2w1, \cdots, 9w1,
    1w9]}$
} \\
\bottomrule
\end{tabular}
\end{adjustbox}
\caption{
Statistical of the vocabularies of different personalities.
}
\label{tab:personality}
\end{table}
We obtain the personality of the characters in six television shows from the Personality Database (PDB) website. 
Each character's personality is determined through crowd-sourced voting on this site. 
We adopt the personality that constitutes the largest proportion of votes as the character's personality.

Taking into consideration that personality is assessed using multiple evaluation criteria, we have annotated the mainstream methods in psychology, namely Myers-Briggs Type Indicator ($\mathtt{MBTI}$), Five Factor Model ($\mathtt{FFM}$), and $\mathtt{Enneagram}$, as shown in Table~\ref{tab:personality}. 
For example, Penny (a character in the television show The Big Bang Theory) possesses ESFP ($\mathtt{MBTI}$), SCUAN ($\mathtt{FFM}$), and 7w8 ($\mathtt{Enneagram}$) personalities, and is an \emph{confident} and \emph{lively} individual. 
More details can be found in the Appendix~\ref{sec:Personality_append}.

\subsection{Dataset Statistics}
\label{section:dataset split}
It consists of 225 distinct individuals, each characterized by their own unique personalities, extracted from six television series. 
Table~\ref{tab:Statistics_of_task1} provides an overview of the statistics pertaining to the PHMRD dataset. 
On average, each segment in the PHMRD dataset has a duration of 74.95 seconds and featured approximately 3.73 characters.
The specific number of samples that conform to the task settings within each television show is presented in Table~\ref{tab:six_tv_shows}.

\section{Method}
We propose a preliminary framework, 
Personality-aware Reasoning Model (\textbf{PRM}),
to validate the effectiveness of personality in our PHMR task. 
The overall structure of \textbf{PMR} is shown in Fig.~\ref{fig:model}. 

\subsection{Multimodal Signal Representation}
Since a multimodal model pre-trained on large scale data can better represent features from different modalities, we utilize a pre-trained model, Merlot Reserve~\citep{merlot}, which is the SOTA of TVQA, for feature extraction.
We employ the notation $\mathcal{V}$, $\mathcal{A}$, $\mathcal{D}$, $\mathcal{P}$, $\mathcal{C}$ to represent the video, audio, dialogue, personality, and choices mentioned in Section~\ref{sec:task}. 

For video modality $\mathcal{V}$, we employ RESERVE-L that utilizes a 24-layer ViT-L/16~\citep{vit} to encode each frame independently.
It obtains $V' \in \mathbb{R}^{n_V \times d}$,
where $n_V$ represents the length of the image frame sequence, and $d$=1024 denotes the feature dimension.

For audio modality $\mathcal{A}$, we utilize an Audio Spectrogram Transformer~\citep{ast} to encode each subsegment independently. 
This is achieved by dividing the audio into intervals of 5 seconds. 
It yields $A' \in \mathbb{R}^{n_A \times d}$,
where $n_A$ represents the number of subsegments after the segmentation process. 

For dialogue $\mathcal{D}$, the context of $\mathcal{D}$ is represented as $D' \in \mathbb{R}^{n_D \times l_D \times d_E}$. 
Here, $n_D$ denotes the number of dialogue utterances, and $l_D$ signify the number of words in a single utterance. 
Specifically, the representation for each word in the utterance is initialized from Merlot Reserve's embedding matrix.

To represent personality $\mathcal{P}$,  we first construct vocabularies corresponding to personality traits based on their discrete distribution.
The personality embedding matrix is randomly initialized and optimized in training. 
We have the option to utilize a single personality trait, such as $\mathtt{[ESFP]}$, or to utilize their collective set, such as $\mathtt{[ESFP; SLUAN; 7w8]}$. 
Ultimately, we obtain the personality feature $P' \in \mathbb{R}^{n_{P} \times l_{P} \times d_E}$ of relevant the characters in the $\mathcal{VC}$. 
Here, $n_P$ denotes the number of relevant characters in $\mathcal{VC}$, and $l_P$ represents length of $P$.

The representation of the multiple choice $\mathcal{C}$ is expressed as $C \in \mathbb{R}^{n_C \times l_C \times d_E}$. 
Here, $n_C$ represents the number of multiple choice options, and $l_C$ signify the number of words in a single choice, respectively. 
Specifically, the representation for each word in the utterance is initialized from Merlot Reserve's embedding matrix. 

\subsection{Personality-aware Multimodal Reasoning}
First, the candidate behaviors are integrated with multimodal and personality features. 
In particular, we concatenate $P$ and $C$ and feed them into Answer-Attention. 
We apply mean-pooling on $P_C$ to produce the final $P^{out}$. 
Similarly, $D$ and $A$ are processed to obtain $D^{out}$ and $A^{out}$, respectively. 

Subsequently, the multimodal features are fused with the personality features.
We first concatenate $P^{out}$ with $D^{out}$ and $A^{out}$, respectively. 
Then, multimodal information is fused to obtain the modal answer. 
Finally, we utilize a softmax layer to derive the final result as follows:
\begin{equation} 
Ans' =[D^{out};P^{out}] + [A^{out};P^{out}],
\end{equation}
\begin{equation} 
Ans=\text{Softmax}(Ans'). 
\end{equation}

In addition to the above method specifically designed for the task, we also modify and adapt several methods from related tasks (TVQA, PMR)
to suit our task.
Details can be found in Sec.~\ref{sec:comparemethods}.

\begin{figure}[htb]
    \centering
    \includegraphics[width=0.5\textwidth]{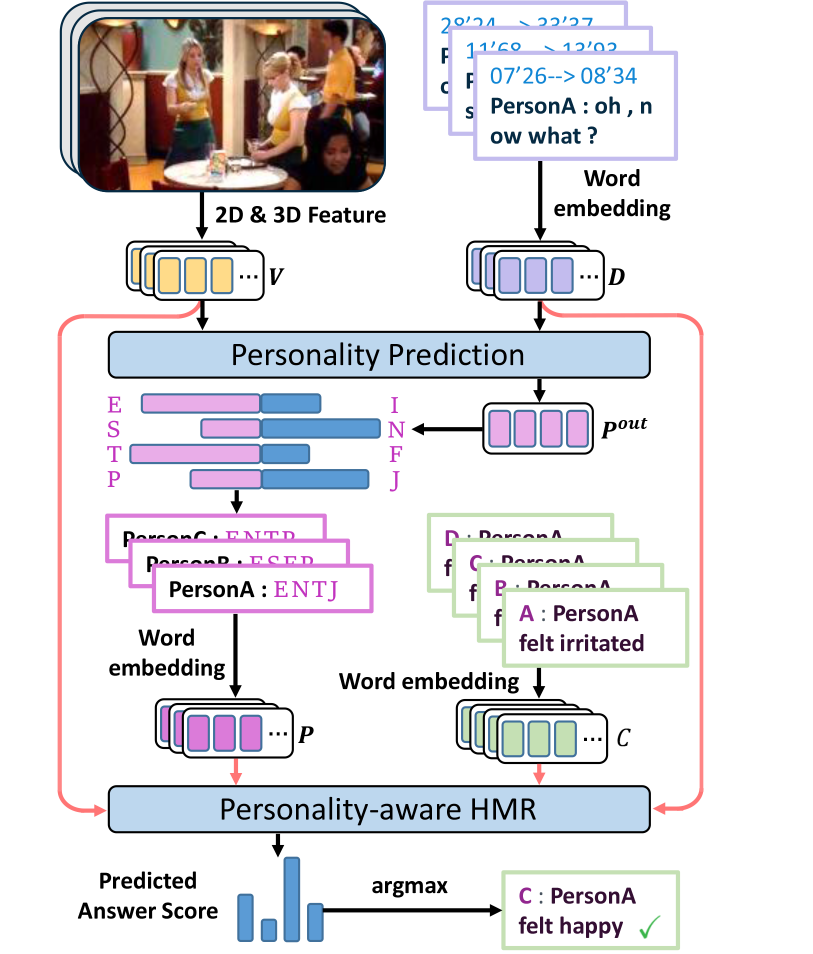}
    \caption{
    PRM$_{pretrain}$ for Personality prediction and Personality-aware HMR.
    }
    \label{fig:task2totask1}
\end{figure}
\section{Extension: Human-centric Multimodal Reasoning with Predicted Personality} 
As previously mentioned, acquiring personality annotations in real life is challenging. 
To overcome this challenge, we propose an extension task called Personality-predicted Human-centric Multimodal Reasoning ($T^2$). 

As shown in Fig.~\ref{fig:task2totask1}, given the availability of multimodal information, we first predict the most probable personality traits $\mathcal{P'}$ of a specific individual in the ${\mathcal{VC}}$. 
We accomplish this through the following process:
\begin{equation}
\mathcal{P'} = {\text{Personality-Prediction}}(\mathcal{D},\, \mathcal{V},\,\mathcal{A}). 
\end{equation}
Second, we substitute the annotated personality $\mathcal{P}$ with the predicted one $\mathcal{P'}$ in $T^1$ and re-evaluate its performance by
\begin{equation} 
\mathcal{B} = {\rm PHMR}(\mathcal{D},\, \mathcal{V},\,\mathcal{A},\, \mathcal{P'}). 
\end{equation}

\label{sec:MPP}
In order to achieve precise predictions of personality traits and effectively utilize these predictions for future forecasting, it is imperative to ensure the complete separation of the MPPD and PHMRD. 
This precautionary step is taken to prevent any data leakage, ensuring that the MPPD remains distinct and independent from the PHMRD. 
To accomplish this, we consider all clips within the TVQA dataset as clips within the MPPD. Subsequently, we exclude any clips that are also present in the PHMRD. 

\begin{table}
\centering
\begin{adjustbox}{max width=0.8\columnwidth}
\resizebox{0.40\textwidth}{!}{
\begin{tabular}{*{5}{c}}
\toprule
{} &{Train} &{Dev}& {Test} & {Total}\\
\midrule
{\# Sample} & {8152} & {1746} &{1747} &{11645}\\
{\# Personality} & {3.29} & {3.25} &{3.27} &{3.28} \\
\bottomrule
\end{tabular}
}
\end{adjustbox}
\caption{Basic statistic of our MPPD dataset.
}
\label{tab:Statistics_of_task2}
\end{table}
Furthermore, we eliminate characters from the dataset that possess insufficient information, such as Person $\mathit{0}$, who only engages in greetings or constantly eats. This process yields a list of characters associated with each clip. 
Once we obtain the corresponding personality traits based on these characters, we replace the names of the individuals with name tags, intentionally disrupting the association between the name and the personality traits. This precautionary measure prevents the model from memorizing the link between specific names and corresponding personality traits. 
The statistical details of the dataset are summarized in Table~\ref{tab:Statistics_of_task2}. 

\section{Experiments}

\subsection{Experimental Settings}
The PHMRD and MPPD dataset is divided into training, validation, and testing sets at a ratio of 14:3:3. 
We use the Accuracy as the primary evaluation metric. 

We train our \textbf{PRM} model on four RTX 3090 GPUs.
For all the models, we train in 20 epochs with Early Stopping. 
We set the learning rate to $8e-5$ for linear layer in feature fusion, $4e-5$ for personality attention in personality feature extraction, and $1e-5$ for the rest, with linear warmup process. 
We report the average performance of four runs. 

\subsection{Compared Methods}
\label{sec:comparemethods}
We establish three rule-based methods, three adapted multimodal reasoning models, a modality separation variant, and two large video models, as compared methods to our \textbf{PRM} model. 

\paragraph{Rule Based Methods.}
\textbf{Random}, \textbf{Longest} and \textbf{Shortest} selects a random, the longest and the shortest option, respectively. 
\paragraph{Adapted Methods.}
\textbf{FVTA-adapt}~\citep{FVTA} concentrates on aligning video and text modalities locally. \textbf{TVQA-adapt}~\citep{tvqa} integrates personality and modality information to acquire features for each modality. \textbf{Reserve-adapt}~\citep{merlot} fuses video with text and audio during pre-training, and combines them with personality features to generate the final prediction results. 
\paragraph{Multimodal Large Language Models.}
\textbf{Chat-UniVi}~\citep{chatunivi} employs a visual model to convert the images into embeddings and transforms the dialogue and personality into prompts. \textbf{Gemini}~\citep{Gemini} use 16 images and prompt with dialogue, personality, and multiple choices. 
\paragraph{Modality Separation method.}
\textbf{\textbf{PRM}$_{vanilla}$} is a variant of \textbf{PRM}.
It generates independent representations for each of the three modalities.
This approach allows us to conduct modality ablation experiments more easily and effectively.
To address the challenges posed by the integration of video modality features into the audio and text modalities as observed in the Merlot Reserve experiment, we also introduce a vanilla method.

\begin{table}
\centering
\begin{adjustbox}{max width=0.8\columnwidth}
\begin{tabular}{*{4}{c}}
\toprule
{\textbf{Model}} &{w/o Per.} &{w/ Per.}& {Imp.}  \\
\midrule
{Shortest} & {16.70} & {-} & {-}  \\
{Random} & {20.00} & {-} & {-}  \\
{Longest} & {\textbf{34.08}} & {-} & {-}  \\
\midrule
{FVTA-adapt} & {33.64} & {34.35} & {0.71}  \\
{TVQA-adapt} & {\textbf{43.37}} & {43.92} & {0.55}  \\
{PRM$_{vanilla}$} & {43.03} & {\textbf{44.14}} & {\textbf{1.11}}  \\
\midrule
{Reserve-adapt} & {52.43} & {53.07} & {0.64}  \\
{\textbf{PRM}} & {\textbf{53.13}} & {\textbf{54.06}} & {\textbf{0.93}}  \\
\bottomrule
\end{tabular}
\end{adjustbox}
\caption{
Accuracy for rule-based, deep learning methods and video large model. Per. represents personality, while Imp. denotes the enhancement achieved by incorporating personality information.
}
\label{tab:task1MainResult}
\end{table}

\subsection{Main Results}
In Table~\ref{tab:task1MainResult}, we report the results of different compared methods in two settings. 
In the supervised setting, compared to TVQA-adapt, although the \textbf{PRM}$_{vanilla}$ model's performance is suboptimal without incorporating personality, it achieves the best performance when personality information is integrated. 
Compared to Reserve-adapt, the \textbf{PRM} model exhibits a significant improvement. 
It highlights that personality information is effective for the task and our model can more effectively utilize personality information. 

\begin{table}
\centering

\begin{adjustbox}{max width=0.8\columnwidth}
\begin{tabular}{*{4}{c}}
\toprule
{\textbf{Model}} &{w/o Per.} &{w/ Per.}& {Imp.}  \\
\midrule
{Chat-UniVi} & {23.50} & {20.61} & {-2.89}  \\
{Gemini} & {33.17} & {34.08} & {0.91}  \\
\bottomrule
\end{tabular}
\end{adjustbox}
\caption{
Accuracy for zero-shot multimodal large language models.
}
\label{tab:zero-shot}
\end{table}

In the zero-shot setting, as shown in Table~\ref{tab:zero-shot}, the existing multimodal large language models face challenges in PHMR and exhibit performance inferior to fine-tuned pretraining models. 
In contrast to Chat-UniVi, Gemini exhibits enhanced performance by incorporating personality. 
This enhancement may be attributed to the inclusion of the correspondence between personality and behavior in Gemini's training corpus, such as the inclusion of task-oriented datasets for personality prediction.

\begin{table}
\centering
\begin{adjustbox}{max width=0.8\columnwidth}
\begin{tabular}{*{3}{lcc}}
\toprule
{\textbf{Personality}} &{Accuracy}& {Imp.}  \\
\midrule
{Baseline} & {53.13} & {-} \\
{w/ $\mathtt{MBTI}$} & {54.03} & {0.90}  \\
{w/ $\mathtt{FFM}$} &  {54.04} & {0.91}  \\
{w/ $\mathtt{Enneagram}$}  & {53.89} & {0.76}  \\
{w/ All} & {\textbf{54.06}} & {\textbf{0.93}}  \\
\bottomrule
\end{tabular}
\end{adjustbox}
\caption{
Accuracy for different personalities for \textbf{PRM}. 
}
\label{tab:ablation_personality}
\end{table}

\subsection{Results of Different Personality Type}

Table~\ref{tab:ablation_personality} displays the outcomes derived from experiments involving different personality types. 
The experimental results demonstrate that the combination of three personalities yields the best performance, followed by $\mathtt{MBTI}$ and $\mathtt{FFM}$, which achieve comparable results. 

The $p$-value obtained from the statistical significance experiment is lower than 5\%, 
signifying that the improvement attributed to the incorporation of personality information is statistically significant. 

\subsection{Results of Different Modalities}
\begin{table}
\centering
\begin{adjustbox}{max width=0.8\columnwidth}
\begin{tabular}{*{4}{c}}
\toprule
{\textbf{Modality}} &{w/o Per.} &{w/ Per.}& {Imp.}  \\
\midrule
{Dialogue} & {51.02} & {51.27} & {0.25}  \\
{Audio} & {52.82} & {53.55} & {0.73}  \\
{All} & {\textbf{53.13}} & {\textbf{54.06}} & {\textbf{0.93}} \\ 
\bottomrule
\end{tabular}\end{adjustbox}
\caption{
Accuracy for different modality for \textbf{PRM}. All represents input of all modality information.
}
\label{tab:ablation_multimodal}
\end{table}

As showed in Table~\ref{tab:ablation_multimodal}, we conduct the experiments under various modality combinations based on \textbf{PRM}. 
In the unimodal settings, audio outperforms dialogue both in terms of base performance and the enhancement contributed by personality. 
The multimodal result is higher than the unimodal one, with a more substantial improvement in personality enhancement.

\begin{table}
\centering
\begin{adjustbox}{max width=0.8\columnwidth}
\begin{tabular}{*{4}{c}}
\toprule
{\textbf{Modality}} &{w/o Per.} &{w/ Per.}& {Imp.}  \\
\midrule
{Dialogue} & {43.08} & {43.37} & {0.29}  \\
{Audio} & {36.59} & {36.70} & {0.11}  \\
{Video} & {43.74} & {44.23} & {0.49}  \\
{All}  & {43.03} & {44.14} & {\textbf{1.11}}  \\
\bottomrule
\end{tabular}\end{adjustbox}
\caption{
Accuracy for different modality for \textbf{PRM}$_{vanilla}$. 
}
\label{tab:ablation_multimodal_cla}
\end{table}
As demonstrated in Table~\ref{tab:ablation_multimodal_cla}, the results for different modalities in \textbf{PRM}$_{vanilla}$ are presented. 
The audio modality exhibits low performance, thus diminishing the overall effectiveness of the vanilla method across all modalities.
Nevertheless, the incorporation of personality information yields the most significant enhancement in the context of multimodality.

\begin{table}
\centering
\begin{adjustbox}{max width=0.8\columnwidth}
\begin{tabular}{lccc}
\toprule
{\textbf{Test Domain}} &{w/o Per.} &{w/ Per.}& {Imp.}  \\
\midrule
{Friends} & {52.92} & {53.40} & {0.48}  \\
{BBT} & {55.57} & {56.23} & {0.66}  \\
{Castle} & {48.44} & {49.34} & {0.90}  \\
{Met} & {56.14} & {59.12} & {\textbf{2.98}}  \\
{House} & {48.68} & {49.49} & {0.81}  \\
{Grey} & {57.65} & {58.85} & {1.20}  \\
\bottomrule
\end{tabular}\end{adjustbox}
\caption{
Accuracy for different TV shows for \textbf{PRM}.
}
\label{tab:ablation_tv}
\end{table}
\subsection{Results on Different Television Shows}
In order to account for the diverse types and quantities of episodes in the six television shows, we divide them into partitions to assess the influence of incorporating personality information.
Given the limited number of available episodes for each TV show, training separate models individually proves to be a challenge.
Therefore, we opt for a trained model on the complete dataset and evaluate its performance on the test sets of these six television shows. The corresponding results can be found in Table~\ref{tab:ablation_tv}.
The results show that the introduction of personality in each type of TV shows leads to a improvement, and that the improvement is greater for TV shows with a relatively small number of clips.

\begin{table}
\centering
\begin{adjustbox}{max width=0.8\columnwidth}
\begin{tabular}{*{4}{c}}
\toprule
{\textbf{Model}}
{} &{HL$_\downarrow$}&{RL$_\downarrow$}&{AP$_\uparrow$}  \\
\midrule
{PRM}  &{25.52} &{25.71} &{77.68}  \\
\bottomrule
\end{tabular}\end{adjustbox}
\caption{
Results of personality prediction. Here, HL means Hamming Loss, and RL means Ranking Loss, AP means Average Precision.
}
\label{tab:task2MainResult}
\end{table}
\begin{table}
\centering
\begin{adjustbox}{max width=0.8\columnwidth}
\begin{tabular}{*{4}{c}}
\toprule
{\textbf{Model}}

&{Annotated}&{Predicted}&{Dif.}  \\
\midrule
{{PRM}}     & {54.03} & {53.82} & {\textbf{-0.21}}  \\
\bottomrule
\end{tabular}\end{adjustbox}
\caption{
Different result between predicted personality retraining and annotated ones.
Dif. indicates the difference between them.
}
\label{tab:task2totask1}
\end{table}
\subsection{Experiments on the Extension Task $T^2$}
We utilize the \textbf{PRM} model to train and predict personality traits on the MPPD dataset. Specifically, we predict the Myers-Briggs Type Indicator (MBTI) and apply the predicted MBTI to $T^1$.
\paragraph{Effect of Personality Prediction} 
In Table~\ref{tab:task2MainResult}, the lower is better for hamming loss and ranking loss, while the greater is better for average precision. 
The result indicates that the predicted personality can serve as a reasonable approximation of the ground truth.

\paragraph{Effect of Personality-Predicted PHMR}
To verify the usefulness of the predicted personality, we substitute the annotated personality with the predicted personality and train \textbf{PRM} for $T^1$ from scratch. 
Table~\ref{tab:task2totask1} illustrates the performance, showcasing a marginal decrease of 0.21\%.
The findings show that multimodal information can be utilized to predict personality to alleviate the shortage of personality annotations in real life. 

\section{Related Work}
\subsection{Multimodal Reasoning}

Multimodal reasoning builds upon unimodal reasoning, with examples such as VCR~\citep{VCR}, VCG~\citep{VCOMET}, and NExT-QA~\citep{Next_qa}, which include both object-centric and human-centric reasoning. 

Object-centric reasoning tasks primarily focus on objects within videos, featuring less content about humans. 
AGQA~\citep{agqa} predicts object relations using video and relation graphs.
Sherlock~\citep{Sherlock} extrapolates missing information based on existing clues in images.
TVQA~\citep{tvqa} selects answers based on video, dialogue, question, and time span, with some questions related to humans.

Human-centric reasoning tasks often feature individuals appearing in a video only once, making it difficult to obtain personality information without numerous samples of the same person. 
Social-IQ~\citep{social_qa} focused on multiple-choice questions using video, dialogue, and behavioral and psychological inquiries.
WHYCAT~\citep{whyact} assessed motivation based on video and descriptions.
PMR~\citep{PMR} selected the optimal response using images and questions.

The aforementioned tasks provide the information about the question, whereas others perform reasoning without this information. 
These tasks usually do not involve individuals's psychological activity. 
VLEP~\citep{VLEP}predicted the next likely action based on video and textual description. 
VAR~\citep{VAR} inferred images and descriptions based on videos and surrounding context, following $\rm \alpha$NLI~\citep{ALI}, a task concentrating solely on the text modality.

\subsection{Personality Computing}

Personality computing~\citep{personality_survey} is a research field that investigates personality through computational techniques utilizing various sources such as text, multimedia, and social networks. 
At present, three primary personality evaluation indicators are employed in research: the Five Factor Model ($\mathtt{FFM}$)~\citep{Bigfive}, Myers-Briggs Type Indicator ($\mathtt{MBTI}$)~\citep{mbti}, and $\mathtt{Enneagram}$~\citep{enneagram}. 
The $\mathtt{FFM}$'s five dimensions describe predictable surface behavior, while $\mathtt{MBTI}$'s four dimensions primarily explain behavior and are closely related to instincts. 
The $\mathtt{Enneagram}$ outlines nine core motivations of individuals, each possessing its own patterns of thinking, feeling, and behaving. 

The work related to personality computing has two main branches: predictive personality and applied personality. 
The evolution of the personality prediction task is presented as follows. 
Early studies~\citep{eifrombook} adopted a simpler binary classification evaluation metric. 
Subsequent studies~\citep{ChaLearn,onestagebigfive,bigfivemultimodal} aimed to establish predictions for the Big Five personality traits. 
In recent years, research efforts~\citep{ridditMBTI,eaclmbti,mbtiprediction} have shifted their focus towards the $\mathtt{MBTI}$. 
There were also some studies~\citep{persemon,pdsm,ehm,unify} that focus on utilizing personality in practical applications.

\section{Conclusion}
In this work, we introduce a new Personality-aware Human-centric Multimodal Reasoning (PHMR) task ($T^1$) and construct a new dataset, PHMRD, based on six television shows. 
The experimental results indicate that incorporating personality information enhances the performance of human-centric multimodal reasoning. 
Moreover, an ablation study reveals that three distinct personality traits contribute to varying degrees of performance improvement. 
To further solve the lack of personality annotation in real-life scenes, we introduce an extended task called Personality-predicted Human-centric Multimodal Reasoning ($T^2$). 
The experimental results show that our method can accurately predict personality, and achieves satisfactory multimodal reasoning performance without relying on personality annotations.

\section*{Limitations}
Personality-aware human-centric multimodal reasoning is a challenging task. 
This work is a preliminary study for this task, which was not defined as a generation task. 

Due to space limitation, Our focus in this work is on the presentation of the task and the dataset. 
The proposed baseline is quite simple and has plenty of space for improvement. 

Personality computing is a relatively broad field, our review of literature may not fully cover the research in this area. 

\section*{Ethics Statement}
We would like to thank~\citet{tvqa}'s valuable work on TVQA. 
The TVQA is licensed under a licence of MIT, which allows commercial using, modification, distribution, and private using the material for any purpose. 
We will also make our PHMRD publicly available later. 
Personality database (PDB) website is an open source, anyone can get information without registration. 
All of the datasets and models are in English, which benefits English speakers more. 
We have employed 2 postgraduates experienced in natural language processing for verify the results of ChatGPT. 
We pay postgraduates around \$7-10 per hour, well above the local average wage, and engage in constructive discussions if they are concerned about the process.

\bibliography{custom}

\appendix

\section{ChatGPT Prompt and Some Samples}
\label{sec:example}

\subsection{Data Filtering}
PROMPT:`` \{What color is the object?\} as unrelated to individuals, while \{Why does person $\mathit{0}$ like this object?\} is regarded as pertinent to human subjects. Is \{QUESTION\} related to human behavior or psychology?''

Similarly, the question ``What kind of object is on person $\mathit{0}$'s leg?'' is considered irrelevant to individuals, whereas the query ``What is person $\mathit{0}$ talking about?'' is deemed relevant to human subjects.

\section{Personality typology}
\label{sec:Personality_append}
\subsection{MBTI}
\begin{figure*}
    \centering
    \includegraphics[width=1.0\textwidth]{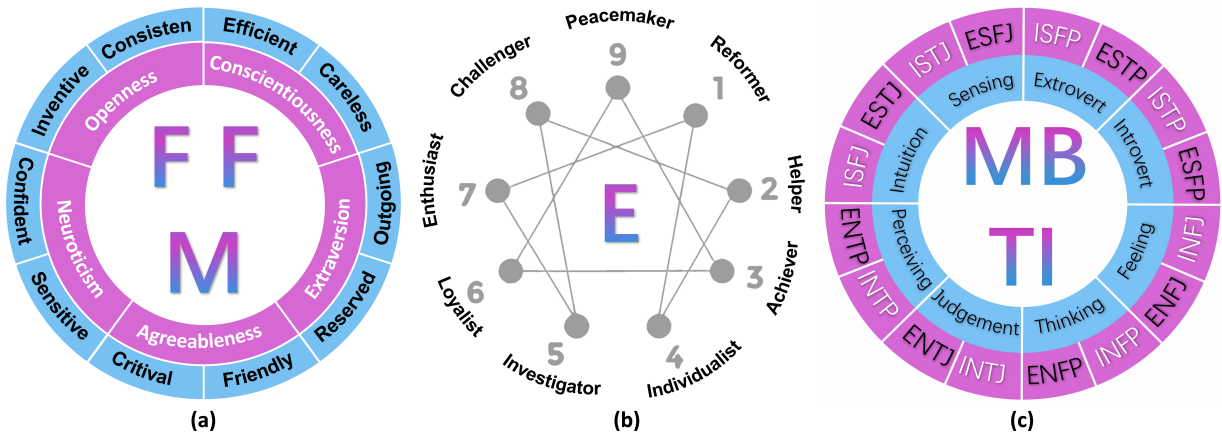}
    \caption{
    three personalities.
    }
    \label{fig:personality}
\end{figure*}

Myers-Briggs Type Indicator ($\mathtt{MBTI}$) is an introspective self-report questionnaire used in personality typology to identify various psychological preferences in how people view the environment and make judgments. 

For evaluation, the $\mathtt{MBTI}$ splits personality into four dimensions (E/I, S/N, T/F, and J/P). 
The four dimensions, in particular, have various emphases, as follows:
\begin{itemize}
 \item Concentrate on attention, Extrovert (E), and Introvert (I) to show the people's primary energy source. 
 \item Concentrate on cognition, Judgment (J), and Perceiving (P) to demonstrate how humans collect information. 
 \item Concentrate on judgment, Intuition (N), and Sensing (S) to demonstrate how individuals make decisions. 
 \item Concentrate on living, Thinking(T), and Feeling(F) to demonstrate how humans behave. 
\end{itemize}
Each dimension includes two opposing labels that may be combined to produce 16 personalities, as seen in Fig.~\ref{fig:personality}(c). 

The 16 personalities have their own characteristics, as shown in 
Table~\ref{tab:mbti_type}. 
\begin{table*}[!ht]
\centering
\resizebox{\textwidth}{!}{
\begin{tabular}{ccp{14cm}}
\toprule
{\textbf{MBTI}} &{Name}  &{Description}\\
\midrule
{\textbf{ISTJ}} & {Inspector} & { These people tend to be reserved yet willful, with a rational outlook on life. They compose their actions carefully and carry them out with methodical purpose.} \\
{\textbf{ISFJ}} & {Protector} & {These people tend to be warm and unassuming in their own steady way. They’re efficient and responsible, giving careful attention to practical details in their daily lives.} \\
{\textbf{INFJ}} & {Counselor} & {They tend to approach life with deep thoughtfulness and imagination. Their inner vision, personal values, and a quiet, principled version of humanism guide them in all things.} \\
{\textbf{INTJ}} & {Mastermind} & {These thoughtful tacticians love perfecting the details of life, applying creativity and rationality to everything they do. Their inner world is often a private, complex one.} \\
{\textbf{ISTP}} & {Crafter} & {They tend to have an individualistic mindset, pursuing goals without needing much external connection. They engage in life with inquisitiveness and personal skill, varying their approach as needed.} \\
{\textbf{ISFP}} & {Composer} & {They tend to have open minds, approaching life, new experiences, and people with grounded warmth. Their ability to stay in the moment helps them uncover exciting potentials.} \\
{\textbf{INFP}} & {Healer} & {These rare personality types tend to be quiet, open-minded, and imaginative, and they apply a caring and creative approach to everything they do.} \\
{\textbf{INTP}} & {Architect} & {These flexible thinkers enjoy taking an unconventional approach to many aspects of life. They often seek out unlikely paths, mixing willingness to experiment with personal creativity.}\\
{\textbf{ESTP}} & {Promoter} & {They tend to be energetic and action-oriented, deftly navigating whatever is in front of them. They love uncovering life’s opportunities, whether socializing with others or in more personal pursuits.} \\
{\textbf{ESFP}} & {Performer} & {These people love vibrant experiences, engaging in life eagerly and taking pleasure in discovering the unknown. They can be very social, often encouraging others into shared activities.} \\
{\textbf{ENFP}} & {Champion} & {These people tend to embrace big ideas and actions that reflect their sense of hope and goodwill toward others. Their vibrant energy can flow in many directions.} \\
{\textbf{ENTP}} & {Inventor} & {They tend to be bold and creative, deconstructing and rebuilding ideas with great mental agility. They pursue their goals vigorously despite any resistance they might encounter.}\\
{\textbf{ESTJ}} & {Supervisor} & {They possess great fortitude, emphatically following their own sensible judgment. They often serve as a stabilizing force among others, able to offer solid direction amid adversity.}\\ 
{\textbf{ESFJ}} & {Provider} & {They are attentive and people-focused, and they enjoy taking part in their social community. Their achievements are guided by decisive values, and they willingly offer guidance to others.}\\
{\textbf{ENFJ}} & {Teacher} & {These warm, forthright types love helping others, and they tend to have strong ideas and values. They back their perspective with the creative energy to achieve their goals.} \\
{\textbf{ENTJ}} & {Fieldmarshal} & {They are decisive people who love momentum and accomplishment. They gather information to construct their creative visions but rarely hesitate for long before acting on them.}\\ 
\bottomrule
\end{tabular}
}
\caption{sixteen types of MBTI and their descriptions.\\
$\,$ \\ 
$\,$ \\ 
$\,$ \\ 
}
\label{tab:mbti_type}
\end{table*}

\subsection{FFM}
Five Factor Model ($\mathtt{FFM}$), also known as Big Five Model was the model to comprehend the relationship between personality and academic behaviors.this model was defined by several independent sets of researchers who used factor analysis of verbal descriptors of human behavior.these researchers began by studying relationships between a large number of verbal descriptors related to personality traits. 

For evaluation, the $\mathtt{FFM}$ splits personality into Five dimensions (I/N, O/U, S/R, A/E, and L/C). 
The Five dimensions, in particular, have various emphases, as follows:
\begin{itemize}
 \item Openness to experience (inventive/curious(I) vs. consistent/cautious(N))
 \item Conscientiousness (efficient/organized(O) vs. extravagant/careless(U))
 \item Extraversion (outgoing/energetic(S) vs. solitary/reserved(R))
 \item Agreeableness (friendly/compassionate(A) vs. critical/rational(E))
 \item Neuroticism (sensitive/nervous(L) vs. resilient/confident(C))
\end{itemize}
Detailed names of the five personality dimensions in polar opposites as shown in Fig.~\ref{fig:personality}(a). 

\begin{table*}[!ht]
\centering
\resizebox{\textwidth}{!}{
\begin{tabular}{cp{14cm}}
\toprule
{\textbf{Enneagram}}   &{Description}\\
\midrule
{\textbf{Perfect}} & {Emphasizes principles, is not easy to compromise, distinguishes between black and white, has high demands on both oneself and others, and pursues perfection.} \\
{\textbf{Helping}} & {Desire to establish a good relationship with others, people-oriented, willing to accommodate others.} \\
{\textbf{Achievement}} &  {Competitive, and measure their own value by achievements, is a workaholic.} \\
{\textbf{Ego}} & {Emotional, afraid of being rejected by others, feeling that others do not understand themselves, doing their own way.} \\
{\textbf{Ideal}} & {likes to think and analyze, has a strong desire for knowledge, but lacks action, and has low requirements for material life.} \\
{\textbf{Doubtful}} &  {Be cautious in doing things, not easy to trust others, have many doubts, like group life, and work hard.} \\
{\textbf{Active}} &  {Optimistic, like novelty, like to follow the trend, don't like pressure.} \\
{\textbf{Leader}} &  {Pursue power, emphasize strength, do not rely on others, and have a sense of justice.}\\
{\textbf{Peaceful}} &  {It takes a long time to make decisions, fears disputes, and prays for harmonious coexistence.} \\
\bottomrule
\end{tabular}
}
\caption{sixteen types of MBTI and their descriptions. }
\label{tab:mbti}
\end{table*}
\subsection{Enneagram}
The $\mathtt{Enneagram}$ personality unfolds according to the nine horns of the ancient totem, revealing nine different inner dynamics that make each person inherently unique as an individual.the nine personality types described by the $\mathtt{Enneagram}$ personality theory are not good or bad; there are recognizable and fundamental differences in the way people with different personality types respond to the world. It is now common knowledge that our personalities are our own, that they filter and interpret what we see and hear, and that the basic principle of Type 9 personality theory is that each of us has one of nine possible "filters" that will keep the blueprint for our lives and the general It is used to protect a certain level within our nature and to form our communication strategy with the outside world. 

Names of the nine personalities as shown in Fig.~\ref{fig:personality}(b). 

\end{document}